\documentclass{article}

\usepackage[main,final]{neurips_2026}

\usepackage[utf8]{inputenc} 
\usepackage[T1]{fontenc}    
\usepackage{natbib}
\usepackage{amsmath}
\usepackage{mathtools}
\usepackage{mathrsfs}
\usepackage{enumerate}
\usepackage{float}
\usepackage{placeins} 
\usepackage{bbding}
\usepackage{booktabs}      
\usepackage{amsfonts}       
\usepackage{nicefrac}       
\usepackage{microtype}      
\usepackage{xcolor}         
\usepackage{multirow}
\usepackage{algorithm}
\usepackage{algpseudocode}
\usepackage{enumitem}
\usepackage{tikz}
\usetikzlibrary{positioning,arrows.meta}

\newcommand{\stitle}[1]{\vspace{1mm}\noindent\textbf{#1}}

\usepackage{url}

\colorlet{revisioncolor}{black}
\newcommand{\rev}[1]{{\color{revisioncolor}#1}}
\makeatletter
\newcommand{\printrevisedbibliography}{%
  \begingroup
  \let\memorigbibitem\bibitem
  \renewcommand{\bibitem}{\@ifnextchar[{\membibopt}{\membibplain}}%
  \bibliographystyle{plainnat}%
  \bibliography{ref}%
  \endgroup
}
\newcommand{\membibcolor}[1]{%
  \def\memtestkey{#1}%
  \def\memtargetkey{zhang2024pddlego}%
  \ifx\memtestkey\memtargetkey\color{revisioncolor}\fi
  \def\memtargetkey{izacard2022contriever}%
  \ifx\memtestkey\memtargetkey\color{revisioncolor}\fi
}
\newcommand{\membibopt}[2][]{\normalcolor\memorigbibitem[#1]{#2}\membibcolor{#2}}
\newcommand{\membibplain}[1]{\normalcolor\memorigbibitem{#1}\membibcolor{#1}}
\makeatother

\title{MemPlan: Memory-Conditioned PDDL Planning for Partially Observable Text Environments}

\author{%
  Tengwei Song$^{1,2,3,\dagger}$ \qquad
  Robert Hoehndorf$^{1}$\thanks{Corresponding author.} \\
  $^{1}$ King Abdullah University of Science and Technology (KAUST), Thuwal, Saudi Arabia \\
  $^{2}$ \AA{}bo Akademi University, Finland \\
  $^{3}$ ELLIS Institute Finland \\[2pt]
  \texttt{songtengwei@gmail.com} \qquad
  \texttt{robert.hoehndorf@kaust.edu.sa}
}

\begin{document}

\maketitle
\begingroup
\renewcommand{\thefootnote}{\fnsymbol{footnote}}
\footnotetext[2]{Work done while at KAUST.}
\endgroup

\begin{abstract}
LLM agents are flexible in language interaction, but using them as long-horizon planners is costly and prone to hallucinated or invalid actions. This paper targets a different role for LLMs: constrained bridges between language and symbolic planning. We propose MemPlan, a memory-conditioned PDDL planning framework for partially observable interactive text environments. MemPlan keeps the PDDL domain fixed and reconstructs the PDDL problem at each step from parsed observations, candidate hypotheses, and interaction memory. Positive graph memory assigns relevance-based action costs to candidate hypotheses for missing objects, locations, or conditions, guiding the planner toward more plausible checks. Negative memory records execution-refuted assumptions and compiles them into exclusion constraints, preventing repeated attempts at candidates already contradicted by feedback. A unified schema-conditioned language--symbol interface, trained offline with supervised fine-tuning, connects text environments to the planning loop through observation-to-fact parsing and symbolic-action-to-command grounding. Experiments on TextWorld, ALFWorld, and Robotouille show that MemPlan improves task completion and step efficiency over language-only and PDDL-aware baselines, while substantially reducing online token cost compared with LLM-planning baselines such as ReAct. The same fine-tuned interface is also reused across TextWorld task types without task-specific fine-tuning. Our code and data are available at: \rev{\url{https://github.com/tewiSong/MemPlan}}.
\end{abstract}

\section{Introduction}

Interactive text environments require agents to solve long-horizon tasks under partial observability and delayed consequences, using natural-language observations and natural-language commands~\citep{cote2018textworld,shridhar2021alfworld,gonzalez2025robotouille}. Recent LLM agents can interpret textual feedback and generate plausible commands, but using LLMs as the primary planner remains costly and unreliable. In interactive settings, they may hallucinate actions, lose track of earlier observations, or fail to maintain a consistent long-horizon plan~\citep{valmeekam2023planning,sun2023adaplanner,zhang2025miragebench}.

Classical planners offer a complementary strength. Given a PDDL domain, a symbolic state, and a goal, planners such as Fast Downward~\citep{helmert2006fast} can search explicitly over action preconditions and effects~\citep{mcdermott1998pddl}. Prior work empirically shows that replacing an LLM planner with a classical planner can save approximately 93.1\% of LLM tokens~\citep{dagan2023dynamic}. The difficulty is that partially observable text environments do not directly provide the complete symbolic state required by a closed-world planner. Static LLM-to-PDDL formalization methods cannot interactively update the planning problem from environment feedback~\citep{liu2023llmp,guan2023worldmodel,gestrin2024nl2plan}.

Existing work addresses this mismatch through two main routes. \textit{Belief and open-world completion methods} make planning possible by first choosing or generating a completed symbolic state~\citep{talamadupula2010integrating,dagan2023dynamic}. This approach works when the completion is reliable, but in text environments a missing object, location, or condition can have many plausible completions. Committing to one completed state before planning can lead to unnecessary search or repeated failed trials. \textit{Iterative PDDL refinement methods} update domain or problem files from new observations and error feedback~\citep{zhang2024pddlego,gong2025pddlego}. This gives the representation more flexibility, but correctness then depends on repeated online formalization, which can introduce malformed PDDL, hallucinated facts, invalid goals, or forgotten observations.

This motivates a different use of symbolic planning under partial observability: a closed PDDL problem is still needed at each planner call, but the agent should avoid committing to a single guessed world state or relying on online LLM rewriting of PDDL files. To this end, we propose MemPlan, a memory-conditioned PDDL planning framework for partially observable text environments. MemPlan treats missing task-relevant state as candidate symbolic hypotheses and compiles interaction memory into the PDDL problem before each planner call. Positive graph memory ranks hypotheses for missing objects, locations, or conditions through action costs, allowing the planner to test more relevant candidates first. Negative memory records failed searches or refuted assumptions as symbolic exclusions, preventing repeated attempts at candidates already contradicted by the environment. A frozen schema-conditioned language--symbol interface connects text observations and commands to this symbolic loop through observation-to-fact parsing and symbolic-action-to-command grounding.

Our contributions are as follows. First, we introduce memory-conditioned PDDL problem construction, enabling a closed-world planner to operate in partially observable text environments without online PDDL rewriting. Second, we propose a positive/negative memory mechanism that ranks unresolved hypotheses with graph-derived costs and excludes execution-refuted candidates from later planning calls. Third, we train an offline schema-conditioned language--symbol interface for parsing observations and grounding symbolic actions, reducing online LLM use to constrained interface calls. We evaluate MemPlan on TextWorld, ALFWorld, and Robotouille, measuring task completion, step efficiency, token cost, and cross-task use of the learned interface.

\section{Related Work}

\stitle{State, memory, and feedback in language agents.}
A large body of work studies how language agents act in interactive environments without an explicit symbolic planner. In text-based games, Keep CALM uses language models to generate action candidates and combines them with reinforcement learning for action selection~\citep{yao2020keep}. More recent work focuses on maintaining internal state over long interactions. QuBE constructs question-based belief states to reduce reasoning derailment under partial observability~\citep{kim2024qube}, while WKM introduces a world-knowledge model with global task knowledge and local dynamic state knowledge for agent planning~\citep{qiao2024wkm}. AriGraph builds semantic and episodic graph memory for LLM agents in interactive text games~\citep{anokhin2025arigraph}. Feedback-driven agents such as AdaPlanner refine language plans online using environment feedback~\citep{sun2023adaplanner}. These methods improve memory, action generation, or feedback use for LLM policies. MemPlan uses memory differently: graph-memory relevance is compiled into PDDL hypothesis-action costs, and execution-refuted assumptions are compiled into symbolic exclusions for a classical planner.

\stitle{LLM-based PDDL formalization and planner interfaces.}
Another line of work connects language models with classical planning. LLM+P translates natural-language planning problems into PDDL, calls a classical planner, and translates the symbolic plan back into natural language~\citep{liu2023llmp}. Guan et al. use LLMs to construct PDDL world models and incorporate corrective feedback from validators and humans~\citep{guan2023worldmodel}. LLM-DP moves toward embodied interaction by using LLM-generated beliefs to sample plausible world states for symbolic planning~\citep{dagan2023dynamic}. PDDLEGO and PDDLEGO+ study partial observability by iteratively formalizing, planning, growing, and refining PDDL representations through interaction~\citep{zhang2024pddlego,gong2025pddlego}. GenPlan studies a different interface: using pretrained LLMs to synthesize generalized planners for PDDL domains~\citep{silver2023genplan}. These works show that LLMs can provide useful interfaces to PDDL planners, but they mainly focus on generating, completing, or refining planning representations. MemPlan assumes a fixed PDDL domain and studies how online interaction memory can be compiled into the PDDL problem through candidate hypotheses, action costs, and exclusions.

\stitle{Open-world and structured embodied planning.}
Talamadupula et al. integrate a closed-world planner with an open-world robot using open-world quantified goals, runtime objects, execution monitoring, replanning, and soft goals~\citep{talamadupula2010integrating}. This work motivates the use of planner-internal placeholders for objects that may be discovered during execution. MemPlan adapts this idea to language-mediated environments, where both observations and executable actions are textual. The agent must therefore parse text observations into symbolic facts, ground symbolic actions back to admissible commands, and update memory from textual feedback. Recent structured embodied planning systems also use intermediate symbolic or graph representations: LLM-REGRESS performs open-world embodied planning with lifted regression and LLM-inferred affordances~\citep{liu2025llmregress}; DELTA constructs planning problems from scene graphs and LLM decompositions~\citep{liu2024delta}; LLM+MAP combines LLM reasoning with PDDL-style multi-agent planning for bimanual manipulation~\citep{chu2025llmmap}; and HIVE uses a PDDL-empowered formal backbone for multimodal instruction execution~\citep{vyas2025hive}. These systems highlight the value of structured representations, while MemPlan focuses on partially observable text environments and memory-conditioned PDDL problem construction.

\section{Method}
\label{sec:method}

\begin{figure}
    \centering
    \includegraphics[width=\linewidth]{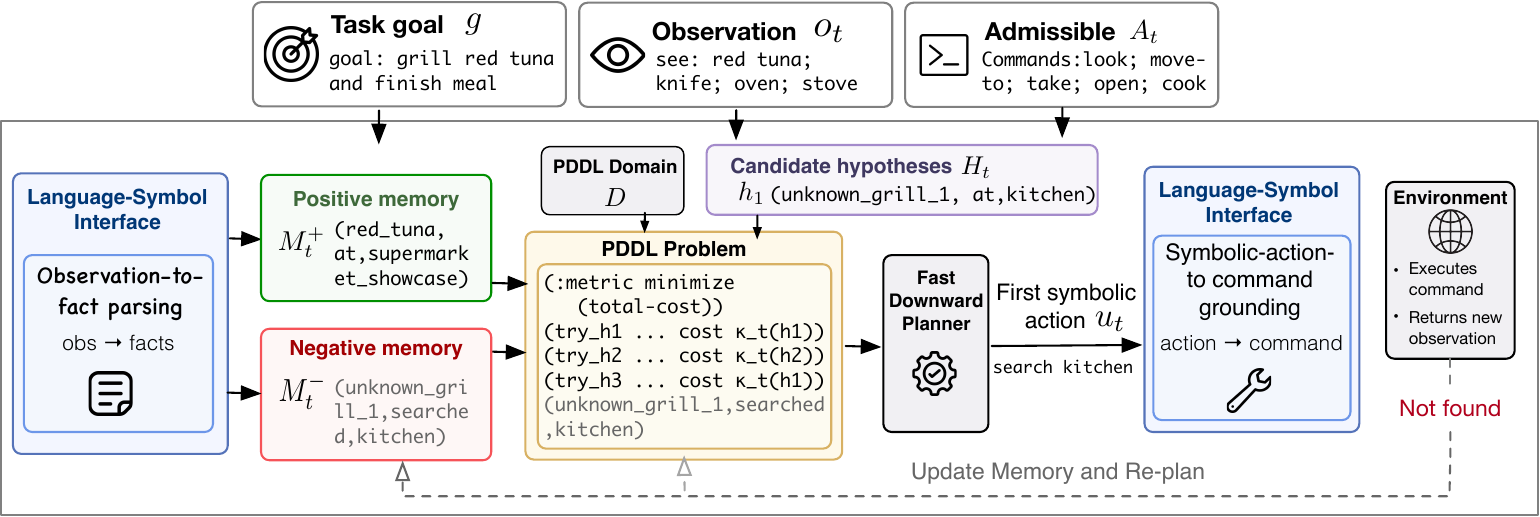}
    \caption{MemPlan online planning loop. Given the task goal, current observation, and admissible commands, the language--symbol interface extracts symbolic facts and grounds planner actions. MemPlan builds a PDDL problem from the fixed domain, candidate hypotheses, positive memory, and negative memory. Environment feedback updates memory and triggers replanning.}
    \label{fig:memplan_overview}
\end{figure}

We study planning in partially observable text environments. At step $t$, the agent receives a task goal $g$, a textual observation $o_t$, and an admissible command set $\mathcal{A}_t$. The agent must select a command $c_t \in \mathcal{A}_t$ without access to the full environment state. MemPlan assumes a fixed PDDL domain $\mathcal{D}$, which specifies object types, predicates, action schemas, preconditions, and effects. The domain is provided to the agent and is not generated or repaired online.

Classical PDDL planners require a closed planning problem, while text environments expose only partial observations. MemPlan addresses this mismatch by rebuilding a closed PDDL problem at each step from parsed facts, candidate hypotheses, and interaction memory. Figure~\ref{fig:memplan_overview} summarizes the online loop. A frozen schema-conditioned language--symbol interface parses observations into symbolic facts, represented internally as typed triples, and grounds symbolic planner actions into executable commands. The planner searches over a PDDL problem constructed from the fixed domain, task goal, parsed facts, candidate hypotheses, positive memory, and negative memory.

Let $F_t$ denote the validated triples extracted from $o_t$, $\mathcal{H}_t$ the candidate hypotheses for unresolved entities required by the goal, $M_t^+$ the positive memory, and $M_t^-$ the negative memory. Let $\mathcal{P}_t$ denote the PDDL problem instance constructed for step $t$. MemPlan builds this problem as
\[
\mathcal{P}_t =
\mathrm{BuildProblem}(\mathcal{D}, g, F_t, \mathcal{H}_t, M_t^+, M_t^-).
\]
The planner returns a symbolic plan $\pi_t=(u_t^{(1)},u_t^{(2)},\ldots)$. MemPlan executes only the first symbolic action $u_t=u_t^{(1)}$, observes the resulting feedback, updates memory, and replans.

\subsection{Offline Schema-Conditioned Language--Symbol Interface}
\label{sec:interfaces}

MemPlan uses one language--symbol interface with two instruction formats. The interface is trained offline by supervised fine-tuning and frozen during evaluation. Online, it only performs observation-to-fact parsing and symbolic-action-to-command grounding. It does not generate plans, repair the PDDL domain, or rewrite PDDL problems.

\paragraph{Observation-to-fact parsing.}
Let $\mathcal{S}_{\mathcal{D}}$ denote the predicate and type schema induced by $\mathcal{D}$, and let $\mathcal{O}_t$ denote the current object glossary. Given $(o_t,\mathcal{S}_{\mathcal{D}},\mathcal{O}_t)$, the parser outputs candidate triples. A triple is accepted into $F_t$ only when its relation appears in $\mathcal{S}_{\mathcal{D}}$, its argument types match the schema, and its entities are grounded in $\mathcal{O}_t$ or introduced as placeholders used only inside planning. The parser extracts local symbolic evidence from the observation; MemPlan constructs the full PDDL problem.

\paragraph{Symbolic-action-to-command grounding.}
Given the first symbolic action $u_t$, the current observation $o_t$, and the admissible command set $\mathcal{A}_t$, the grounder scores each command $c \in \mathcal{A}_t$ with $s_\theta(u_t,o_t,c)$. Let $\mathcal{A}_t(u_t)\subseteq\mathcal{A}_t$ denote admissible commands that can realize $u_t$. When $\mathcal{A}_t(u_t)$ is nonempty, MemPlan selects
\[
c_t=\arg\max_{c\in\mathcal{A}_t(u_t)} s_\theta(u_t,o_t,c).
\]
When $\mathcal{A}_t(u_t)$ is empty, MemPlan selects an admissible observation command, updates memory only from environment feedback, and replans.

\subsection{Memory and Candidate Hypotheses}
\label{sec:memory_and_hypotheses}

Let $\mathcal{E}_t$ be the current entity set. MemPlan stores memory as typed triples over entities. Positive memory is
$M_t^+ \subseteq \mathcal{E}_t \times \mathcal{R}_{\mathcal{D}} \times \mathcal{E}_t$,
where $\mathcal{R}_{\mathcal{D}}$ is the set of positive relations defined in the PDDL domain. These triples are supported by observations or successful execution.

Negative memory is
$M_t^- \subseteq \mathcal{E}_t \times \mathcal{R}_{\mathcal{D}}^- \times \mathcal{E}_t$,
where $\mathcal{R}_{\mathcal{D}}^-$ contains relations that restrict future planning, such as \texttt{searched} and \texttt{not\_at}. Negative triples are added after targeted checks whose feedback indicates that a candidate was searched and the required entity was not found.

When the goal requires an entity that has not been resolved, MemPlan introduces a typed placeholder $z$ used only inside planning. A candidate hypothesis is a triple $h=(z,r,v)$, where $r$ is a relation and $v$ is a candidate entity, container, location, state, or condition. The candidate set $\mathcal{H}_t$ is generated from observed entities, memory graph neighborhoods, recently visited areas, and values compatible with the current goal. Once feedback resolves the placeholder to a concrete entity, later PDDL problems use the concrete entity.

This placeholder mechanism is inspired by open-world planning with runtime objects and execution monitoring~\citep{talamadupula2010integrating}. MemPlan adapts it to text environments by pairing placeholders with learned parsing, learned grounding, and memory-based hypothesis testing.

\subsection{Memory-Conditioned PDDL Construction}
\label{sec:pddl_construction}

\textsc{BuildProblem} converts the current interaction history into the closed PDDL problem $\mathcal{P}_t$. It writes grounded entities and unresolved placeholders into the object list; places unresolved placeholders in a virtual location such as \texttt{loc\_limbo}; writes validated facts $F_t$ and schema-compatible positive-memory triples $M_t^+$ into the initial state; inserts hypothesis-checking actions for candidates in $\mathcal{H}_t$; and compiles negative memory $M_t^-$ into exclusions. The goal condition is derived from $g$, while all action schemas remain fixed in $\mathcal{D}$.

\paragraph{Positive memory as hypothesis costs.}
For each candidate $h=(z,r,v)\in\mathcal{H}_t$, MemPlan computes a relevance score $\mathrm{Score}_t(h)\in[0,1]$ using \rev{frozen multilingual Contriever (mContriever) embeddings~\citep{izacard2022contriever}}. The retrieval query is linearized from the task goal $g$ and the unresolved placeholder $z$. The candidate document is linearized from the candidate triple $h$ and the local neighborhood of $v$ in the positive-memory graph induced by $M_t^+$. \rev{Here $M_t^+$ is a typed directed multigraph: observed or resolved entities are nodes, and each validated triple $(u,r,v)$ creates a relation-labeled edge $u\xrightarrow{r}v$. Candidate documents use the one-hop neighborhood of $v$.} The relevance score is derived from the cosine similarity between the query and candidate-document embeddings. \rev{We use the pretrained \texttt{facebook/mcontriever} checkpoint and apply min--max normalization to the cosine similarities over the candidate set.} It is used only to rank hypothesis-checking actions.
This use of dense graph-memory retrieval is related to AriGraph~\citep{anokhin2025arigraph}, but the retrieved relevance is used differently: MemPlan compiles it into PDDL action costs rather than using it as context for LLM decision-making.

MemPlan converts this score into a PDDL action cost. Physical actions, such as movement, pickup, and manipulation, receive the base cost $\kappa_{\mathrm{base}}$. Hypothesis-checking actions test unresolved hypotheses before the planner commits to downstream physical actions. Their cost is
\[
\kappa_t(h)
=
\kappa_{\mathrm{base}}
+
\left\lceil
\lambda \cdot \max(0,\tau-\mathrm{Score}_t(h))
\right\rceil ,
\]
where $\kappa_t(h)$ is the cost of checking hypothesis $h$ at step $t$, $\tau$ is a relevance threshold, and $\lambda$ controls the penalty scale. Higher-relevance hypotheses receive lower costs. The generated PDDL problem includes \texttt{(:metric minimize (total-cost))}, so the planner is guided toward hypotheses better supported by positive memory. Since our planner configuration is satisficing, costs bias the returned plan without guaranteeing a globally minimum-cost plan. \rev{We fix $\kappa_{\mathrm{base}}=1$, $\tau=0.50$, and $\lambda=10$. A nine-setting sensitivity analysis is provided in Appendix~\ref{app:sensitivity}.}

\paragraph{Negative memory as exclusions.}
When execution feedback refutes a candidate, such as a targeted search returning \textit{not found}, MemPlan records the refutation in $M_t^-$. During the next call to \textsc{BuildProblem}, the refuted candidate is omitted or made inapplicable through exclusion predicates, such as \texttt{searched} or \texttt{not\_at}. Thus $M_t^+$ controls which hypotheses are tried earlier, while $M_t^-$ prevents the planner from repeating hypotheses already contradicted by the environment.

This construction differs from iterative PDDL rewriting methods. The PDDL domain remains fixed, and the LLM does not rewrite the domain or problem online. MemPlan rebuilds the problem by injecting placeholders, costing hypotheses with positive graph memory, and excluding refuted hypotheses with negative memory.

\begin{algorithm}[t]
\caption{MemPlan online planning loop}
\label{alg:memplan}
\small
\begin{algorithmic}[1]
\Require Fixed domain $\mathcal{D}$, goal $g$, initial observation $o_0$, frozen interface, planner $\textsc{Plan}$
\State Initialize $M_0^+ \gets \emptyset$, $M_0^- \gets \emptyset$
\For{$t=0,1,2,\ldots$ until termination}
    \State Receive $o_t$ and admissible command set $\mathcal{A}_t$
    \State Parse and validate triples $F_t$ from $(o_t,\mathcal{S}_{\mathcal{D}},\mathcal{O}_t)$
    \State Add supported triples in $F_t$ to $M_t^+$
    \State Generate candidate hypotheses $\mathcal{H}_t$ for unresolved entities required by $g$
    \State Compute $\mathrm{Score}_t(h)$ and $\kappa_t(h)$ for each $h\in\mathcal{H}_t$
    \State Remove candidates excluded by $M_t^-$
    \State Build $\mathcal{P}_t=\mathrm{BuildProblem}(\mathcal{D},g,F_t,\mathcal{H}_t,M_t^+,M_t^-)$
    \State Compute $\pi_t \gets \textsc{Plan}(\mathcal{D},\mathcal{P}_t)$
    \If{$\pi_t$ is empty}
        \State Choose an admissible observation command $c_t \in \mathcal{A}_t$
    \Else
        \State Let $u_t \gets \pi_t[1]$
        \State Ground $u_t$ to $c_t\in\mathcal{A}_t$; use an observation command if no admissible command realizes $u_t$
    \EndIf
    \State Execute $c_t$ and receive feedback and the next observation
    \State Update $M_{t+1}^+$ with supported triples and resolved placeholders
    \State Update $M_{t+1}^-$ with searched or refuted triples from failed targeted checks
\EndFor
\end{algorithmic}
\end{algorithm}

\subsection{Online Planning and Memory Update}
\label{sec:online_loop}

Algorithm~\ref{alg:memplan} summarizes the online loop. MemPlan executes only the first grounded action of each plan, updates $M_t^+$ with supported observations and resolved placeholders, updates $M_t^-$ with searched or refuted hypotheses, and replans after feedback.

\section{Experiments}
\label{sec:experiments}

We evaluate MemPlan on interactive text environments with partial observability. The experiments address four questions.
\textbf{Q1:} Does MemPlan improve task completion and step efficiency over language-only and PDDL-aware baselines, and how does its online LLM token cost compare with LLM-planning baselines?
\textbf{Q2:} How does memory-conditioned PDDL planning compare with memory-augmented LLM agents?
\textbf{Q3:} Can MemPlan handle synchronous and asynchronous long-horizon planning in Robotouille?
\textbf{Q4:} Which components contribute to the observed gains?

\subsection{Experimental Setup}
\label{sec:exp_setup}

\stitle{Benchmarks.}
We evaluate MemPlan on TextWorld~\citep{cote2018textworld}, ALFWorld~\citep{shridhar2021alfworld}, and Robotouille~\citep{gonzalez2025robotouille}. ALFWorld is our main end-to-end benchmark; TextWorld is used for memory-agent comparison, cross-task interface use, and ablations; Robotouille tests synchronous and asynchronous long-horizon planning. We evaluate on 134 ALFWorld episodes, 7 TextWorld task instances, and 151 Robotouille task-seed runs. For Robotouille, success is first averaged over seeds within each task configuration and then macro-averaged across task configurations. Full split details, interaction budgets, metric definitions, and statistical reporting are provided in Appendix~\ref{app:experimental_details}.

\stitle{PDDL domains.}
MemPlan uses one fixed PDDL domain per environment. We use the ALFWorld-released ALFRED domain and construct Robotouille domains from public domain metadata. TextWorld does not provide PDDL domains for the evaluated task families, so we generate its domains offline before evaluation. GPT-5.1 drafts each domain from environment-level task rules and action descriptions, \textit{without} access to test-instance maps, object placements, oracle trajectories, interaction logs, or evaluation feedback. Validator and planner error messages are used for one offline repair round when needed. In our TextWorld protocol, $5/7$ domains ($71.4\%$) are valid on the first attempt, and the remaining two are repaired successfully after one offline feedback round. The repaired domains are fixed for evaluation; MemPlan only rebuilds instance-level problem files online.

\stitle{MemPlan implementation.}
Given the fixed domain for each environment, MemPlan constructs instance-level PDDL problems online from the current observation, task goal, positive memory, negative memory, and candidate hypotheses. Symbolic planning is performed by Fast Downward~\citep{helmert2006fast}.
The language--symbol interface uses Qwen2.5-7B as the backbone. We supervise fine-tuning one unified interface on TextWorld cook and hunt tasks, using 17,000 observation-to-fact examples and 9,600 symbolic-action-to-command examples, then freeze it for all evaluations. ALFWorld and Robotouille interface splits are used only for post-training diagnostic validation; they are not used for supervised fine-tuning, checkpoint selection, or hyperparameter tuning.

\stitle{Baselines and metrics.}
We compare against language-only baselines, PDDL-aware baselines, and TextWorld memory-agent baselines. We report task success, environment steps, and online LLM token usage; metric definitions and evaluation protocols are detailed in Appendix~\ref{app:experimental_details}.

\subsection{Q1: Main Results on ALFWorld}
\label{sec:main_results}

\rev{Table~\ref{tab:alfworld_main} examines performance across the six ALFWorld task types under the same 134-episode protocol, together with episode length and online LLM tokens. Per-task counts and statistical details are reported in Appendix~\ref{app:statistical_reporting}.}

\begin{table}[H]
\centering
\caption{ALFWorld task completion, episode length, and online LLM token usage.}
\label{tab:alfworld_main}
\small
\resizebox{\linewidth}{!}{
\begin{tabular}{lccccccccc}
\toprule
Model & clean & cool & examine & heat & put & puttwo & overall $\uparrow$ & length $\downarrow$ & tokens $\downarrow$ \\
\midrule
LLM-DP & 0.94 & \rev{\textbf{1.00}} & \rev{\textbf{1.00}} & 0.87 & \rev{\textbf{1.00}} & 0.94 & 0.96 & 13.16 & \rev{\textbf{633k}} \\
GPT-4o & 0.13 & 0.19 & 0.39 & 0.09 & 0.38 & 0.12 & 0.21 & 127.18 & 6.79M \\
GPT-4o + Domain & 0.90 & 0.90 & 0.83 & 0.91 & 0.83 & 0.71 & 0.86 & 34.69 & 13.09M \\
ReAct (GPT-4o) & 0.74 & 0.90 & 0.78 & 0.87 & 0.88 & 0.82 & 0.83 & 38.18 & 5.07M \\
ReAct (LLM-DP) & 0.35 & 0.90 & 0.33 & 0.65 & 0.71 & 0.29 & 0.54 & 18.69 & 9.16M \\
MemPlan & \textbf{1.00} & \textbf{1.00} & \textbf{1.00} & \textbf{1.00} & \textbf{1.00} & \textbf{1.00} & \textbf{1.00} & \textbf{6.91} & 906k \\
\bottomrule
\end{tabular}
}
\end{table}

\rev{Across the task types, giving GPT-4o domain descriptions improves completion but still leaves substantially longer trajectories than MemPlan. LLM-DP is more economical in online tokens, whereas MemPlan requires fewer environment steps. These differences illustrate the benefit of using the domain for explicit planning rather than only as prompt context.}

\rev{To test whether the efficiency advantage persists against a stronger language-model agent and iterative PDDL refinement, Table~\ref{tab:alfworld_comparison} compares aggregate results on the same episodes. PDDLEGO+-Native retains its original fine-grained repair feedback; PDDLEGO+-Shared receives the standard textual feedback available to MemPlan.}

\begin{table}[H]
\centering
\caption{\rev{ALFWorld comparison with frontier LLM and PDDL-aware baselines (134 episodes).}}
\label{tab:alfworld_comparison}
\small
\begin{tabular}{lccc}
\toprule
\rev{Method} & \rev{Success (\%) $\uparrow$} & \rev{Steps $\downarrow$} & \rev{Tokens $\downarrow$} \\
\midrule
\rev{ReAct (GPT-5.1)} & \rev{98.5} & \rev{10.92} & \rev{2.74M} \\
\rev{PDDLEGO+-Native} & \rev{94.8} & \rev{14.32} & \rev{2.31M} \\
\rev{PDDLEGO+-Shared} & \rev{92.5} & \rev{15.47} & \rev{1.82M} \\
\rev{LLM-DP} & \rev{96.0} & \rev{13.16} & \rev{\textbf{0.633M}} \\
\rev{MemPlan} & \rev{\textbf{100}} & \rev{\textbf{6.91}} & \rev{0.906M} \\
\bottomrule
\end{tabular}
\end{table}

\rev{Because frontier ReAct approaches ceiling-level completion on ALFWorld, the more informative differences are in interaction steps and online tokens. MemPlan produces shorter trajectories than the compared methods and uses fewer tokens than GPT-5.1 ReAct and both PDDLEGO+ variants. The PDDLEGO+ comparison remains favorable to MemPlan even under Native's more informative feedback; LLM-DP, however, retains the lowest online token cost.}

\subsection{Q2: TextWorld Memory-Agent Comparison}
\label{sec:textworld_memory}

\begin{figure}[t]
\centering
\includegraphics[width=0.95\linewidth]{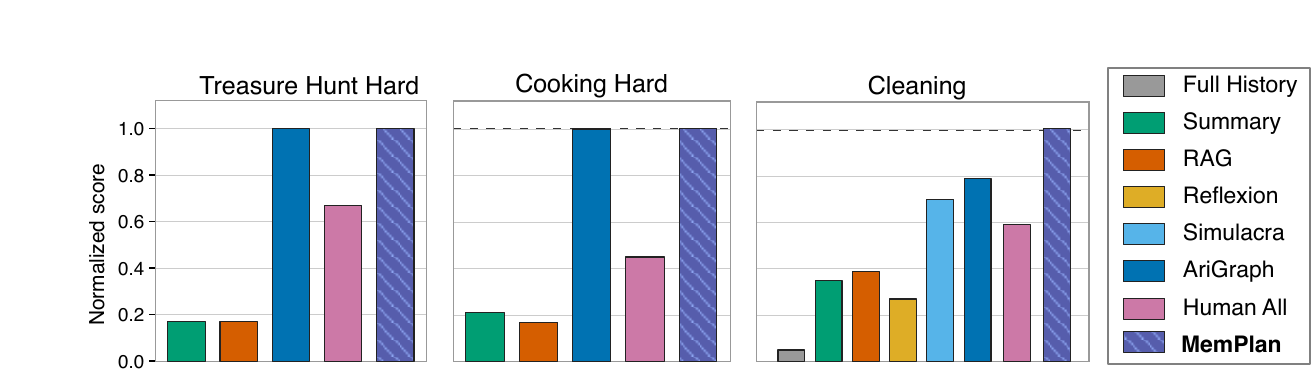}
\caption{TextWorld normalized scores against memory-augmented LLM agents.}
\label{fig:textworld_main}
\end{figure}

Figure~\ref{fig:textworld_main} compares MemPlan with TextWorld memory-agent baselines from AriGraph~\citep{anokhin2025arigraph}. These baselines use the same LLM planning and decision-making backbone but replace the memory module: \textsc{Full History} keeps the complete observation--action history, \textsc{Summary} maintains an iterative summary, \textsc{RAG} retrieves similar past memories, \textsc{Reflexion} uses trial-level verbal reflection, \textsc{Simulacra} scores memories by recency, importance, and relevance, and \textsc{AriGraph} uses a semantic--episodic memory graph. \textsc{Human All} denotes average human performance.

MemPlan uses memory differently from these LLM-memory baselines. Positive memory shapes PDDL hypothesis costs, while negative memory excludes assumptions already contradicted by execution feedback. In the reported TextWorld comparison, MemPlan obtains the strongest normalized-score performance among the compared memory-agent methods in Figure~\ref{fig:textworld_main}. This directly answers Q2: memory-conditioned PDDL planning compares favorably with memory-augmented LLM agents on these TextWorld tasks. The result suggests that memory can be more effective when compiled into planning constraints and costs than when used only as additional LLM context.

Because this comparison uses 7 task instances, we interpret the result as evidence on the reported TextWorld setting rather than as a broad claim over all text-game distributions. Per-instance details and grouping information are reported in Appendix~\ref{app:experimental_details}.

\subsection{Q3: Synchronous and Asynchronous Planning on Robotouille}
\label{sec:robotouille_results}

Table~\ref{tab:robotouille_main} compares synchronous and asynchronous Robotouille tasks, with success macro-averaged over task configurations after averaging across seeds. \rev{MemPlan's lead over frontier ReAct and LLM-DP is more pronounced in the asynchronous setting. This is consistent with feedback-driven replanning being useful when action effects are delayed and previous observations become stale: MemPlan reconstructs the planning problem after each step, incorporates supported facts, and excludes refuted hypotheses.}

\begin{table}[h]
\centering
\caption{Robotouille task success in synchronous and asynchronous settings. Values are macro-averaged over task configurations after averaging over seeds within each configuration.}
\label{tab:robotouille_main}
\small
\begin{tabular}{lcc}
\toprule
Model & Synchronous (\%) $\uparrow$ & Asynchronous (\%) $\uparrow$ \\
\midrule
ReAct (GPT-4o) & 47.0 & 11.0 \\
\rev{ReAct (GPT-5.1)} & \rev{63.6} & \rev{24.7} \\
\rev{LLM-DP} & \rev{57.9} & \rev{18.5} \\
ReAct (GPT-4o-mini) & 11.0 & 0.0 \\
ReAct (Gemini-1.5-Flash) & 0.0 & 0.0 \\
ReAct (Claude-3-Haiku) & 2.0 & 0.0 \\
MemPlan & \textbf{71.2} & \textbf{41.3} \\
\bottomrule
\end{tabular}
\end{table}

\FloatBarrier
\subsection{Q4: Ablation Study}
\label{sec:ablations}

Table~\ref{tab:ablations} evaluates six ablations on the seven TextWorld task instances from Figure~\ref{fig:textworld_main}, with three seeds per instance. We report normalized score, executed steps, repeated hypothesis checks, terminal planner dead ends, and invalid actions. A terminal planner dead end means that the PDDL problem has no usable next symbolic action and the episode terminates. The ablations isolate graph-memory costs, negative memory, placeholders, observation parsing, action grounding, and the full learned interface. Protocol details are provided in Appendix~\ref{app:ablation_details}.

\begin{table}[t]
\centering
\caption{TextWorld ablations. Values are means over 21 runs. Step count is omitted for terminal dead-end variants, where shorter length does not imply higher efficiency.}
\label{tab:ablations}
\small
\resizebox{\linewidth}{!}{
\begin{tabular}{lccccc}
\toprule
Model & Norm. score $\uparrow$ & Steps $\downarrow$ & Rep. checks / run $\downarrow$ & Term. dead ends / run $\downarrow$ & Invalid acts. / run $\downarrow$ \\
\midrule
MemPlan & \textbf{1.00} & \textbf{31.0} & \textbf{0.0} & \textbf{0.0} & \textbf{0.0} \\
Uniform-cost hypotheses & 1.00 & 38.0 & 1.0 & 0.0 & 0.0 \\
w/o negative memory & 0.64 & 60.0 & 12.0 & 0.0 & 1.0 \\
w/o placeholders & 0.27 & -- & 0.0 & 1.0 & 0.0 \\
w/o observation parser & 0.55 & 60.0 & 3.0 & 0.0 & 4.0 \\
w/o action grounder & 0.45 & 60.0 & 1.0 & 0.0 & 9.0 \\
w/o learned interface & 0.36 & 60.0 & 4.0 & 0.0 & 12.0 \\
\bottomrule
\end{tabular}
}
\end{table}

The ablations show three failure modes. Removing placeholders causes structural planning failure because required but unobserved entities cannot be represented in the PDDL problem. Removing negative memory mainly hurts recovery: the agent can still plan, but it revisits hypotheses already contradicted by feedback.

Interface ablations mainly affect execution reliability. Removing the parser weakens the symbolic state, while removing the grounder increases invalid commands. Removing both gives the weakest interface variant. Uniform-cost hypotheses preserve feasibility but require more interaction steps, indicating that graph-memory costs mainly improve search efficiency.

\subsection{Efficiency Analysis}
\label{sec:efficiency}

Table~\ref{tab:efficiency} reports ALFWorld efficiency in terms of environment steps, planner calls, LLM calls, online tokens, and invalid-action rate. Invalid-action rate is the fraction of executed commands rejected by the environment. MemPlan replans after each executed action, so planner calls follow episode length but do not consume LLM tokens. Its online LLM calls are limited to observation parsing and command grounding.

\begin{table}[h]
\centering
\caption{ALFWorld online efficiency breakdown. Invalid-action rate is the fraction of executed commands rejected by the environment.}
\label{tab:efficiency}
\small
\resizebox{0.9\linewidth}{!}{
\begin{tabular}{lccccc}
\toprule
Model & Steps $\downarrow$ & Planner calls $\downarrow$ & LLM calls $\downarrow$ & Tokens / episode $\downarrow$ & Invalid action rate $\downarrow$ \\
\midrule
ReAct (GPT-4o) & 38.18 & -- & 39.78 & 37.9k & 0.03 \\
LLM-DP & 13.16 & 13.16 & 13.16 & 4.7k & 0.01 \\
MemPlan & 6.91 & 6.91 & 16.87 & 6.9k & 0.00 \\
\bottomrule
\end{tabular}
}
\end{table}

MemPlan uses more online tokens than LLM-DP, but achieves shorter episodes and higher task completion in Table~\ref{tab:alfworld_main}. Detailed token-accounting rules are provided in Appendix~\ref{app:token_accounting}.

\subsection{Case Study: Recovering from a Missing Cooking Device}
\label{sec:case_study}

\begin{figure}[ht]
    \centering
\begingroup
\renewcommand{\ttdefault}{lmtt}
\definecolor{casegrey}{RGB}{246,248,250}
\definecolor{caseblue}{RGB}{239,246,255}
\definecolor{casegreen}{RGB}{236,253,245}
\definecolor{caseorange}{RGB}{255,247,237}
\definecolor{casered}{RGB}{254,242,242}
\definecolor{casepillred}{RGB}{239,68,68}
\definecolor{casepillgreen}{RGB}{16,185,129}
\setlength{\fboxsep}{2pt}
\setlength{\fboxrule}{.35pt}
\newcommand{\csfont}{\sffamily\fontsize{5.8}{7.0}\selectfont}
\newcommand{\cscode}{\ttfamily\fontsize{5.3}{6.5}\selectfont}
\newcommand{\csheading}[1]{{\sffamily\bfseries\fontsize{6.7}{8.0}\selectfont #1}\par\vspace{4pt}}
\newcommand{\casebox}[3]{%
  \noindent\fcolorbox{#1!55!black}{#1}{\begin{minipage}{\dimexpr\linewidth-2\fboxsep-2\fboxrule\relax}%
  \raggedright\csfont {\bfseries #2}\par\vspace{1.5pt}#3\end{minipage}}\par\vspace{4pt}}
\pgfmathsetlengthmacro{\csStep}{.20*\linewidth}
\pgfmathsetlengthmacro{\csInner}{.20*\linewidth-16pt}
\pgfmathsetlengthmacro{\csPillWidth}{.76*\csInner}
\pgfmathsetlengthmacro{\csLegendWidth}{5*\csStep-5pt}
\begin{tikzpicture}[x=1pt,y=1pt,
 panel/.style={draw=black!25,inner sep=3pt,anchor=north west,align=left},
 flow/.style={-{Stealth[length=3pt]},line width=.5pt}]
\node[panel] (A) at (0,0) {\begin{minipage}[t][143pt][t]{\csInner}
\csheading{A. Parse\\ observation}
\casebox{casegrey}{Observation}{Kitchen: fridge, oven,\\ stove, knife\par orange bell pepper\\ in fridge\par yellow potato on\\ counter}
\casebox{caseblue}{Parsed symbolic facts}{{\cscode
in(orange\_bell\_\\pepper, fridge)\par
in(yellow\_potato,\\ counter)\par
in(knife, counter)\par
is\_oven(oven\_0),\\ is\_stove(stove\_0)}}
\end{minipage}};
\node[panel] (B) at (\csStep,0) {\begin{minipage}[t][143pt][t]{\csInner}
\renewcommand{\csfont}{\sffamily\fontsize{5.9}{7.1}\selectfont}
\renewcommand{\cscode}{\ttfamily\fontsize{5.6}{6.6}\selectfont}
\csheading{B. Plan and\\ execute}
\casebox{caseblue}{Symbolic plan fragment}{{\cscode
 take bell pepper\\ from fridge\par take knife\par take yellow potato\par cook / chop /\\ add to meal}}
\casebox{casegreen}{Grounded commands}{{\cscode
 take orange bell\\ pepper from fridge\par cook yellow potato\\ with stove\par cook orange bell\\ pepper with oven}}
\end{minipage}};
\node[panel] (C) at (2*\csStep,0) {\begin{minipage}[t][143pt][t]{\csInner}
\renewcommand{\csfont}{\sffamily\fontsize{6.3}{7.8}\selectfont}
\renewcommand{\cscode}{\ttfamily\fontsize{6.2}{7.6}\selectfont}
\csheading{C. Placeholder\\ hypothesis}
\casebox{caseorange}{Unresolved device}{{\cscode assume\_item\_\\found(...)\par unknown\_grill\_1\\ (...)\par}\csfont Planner inserts a\\ placeholder for a\\ required grilling\\ device.}
\casebox{caseblue}{Next symbolic action}{{\cscode cook\_grill(\\ red\_tuna,\\ unknown\_grill\_1)}}
\end{minipage}};
\node[panel] (D) at (3*\csStep,0) {\begin{minipage}[t][143pt][t]{\csInner}
\csheading{D. Grounding\\ failure}
\casebox{casered}{Grounder output}{{\cscode cook\_grill(...\\ unknown\_grill\_1)\par} $\rightarrow$ \texttt{look}\par Diagnosis:\\ \texttt{WRONG\_LOCATION}}
\casebox{casegrey}{Recovery}{execute command:\\ \texttt{look}\par does not treat the\\ unrealized symbolic\\ action as executed;\\ force replanning}

\end{minipage}};
\node[panel] (E) at (4*\csStep,0) {\begin{minipage}[t][143pt][t]{\csInner}
\csheading{E. Replan and\\ finish}
\casebox{caseblue}{Replanned suffix}{{\cscode
 dice(red\_tuna,\\ knife)\par cook\_grill(\\ red\_tuna, bbq\_0)\par add\_ingredient\_\\to\_meal\par finish\_meal, eat}}
\vspace{-3pt}\casebox{casegreen}{Grounded command}{cook red tuna\\ with BBQ\par prepare meal\\ $\rightarrow$ eat meal\par Final score: 14 / 14}

\end{minipage}};
\draw[flow] (A.east) -- (B.west |- A.east);
\draw[flow] (B.east) -- (C.west |- A.east);
\draw[flow] (C.east) -- (D.west |- A.east);
\draw[flow] (D.east |- A.east) -- (E.west |- A.east);
\node[anchor=south,fill=casepillred,rounded corners=7pt,
 minimum width=\csPillWidth,minimum height=13pt,yshift=4pt,
 inner sep=1pt,text=white,font=\sffamily\bfseries\fontsize{6.5}{7.5}\selectfont]
 at (D.south) {Replanning};
\node[anchor=south,fill=casepillgreen,rounded corners=7pt,
 minimum width=\csPillWidth,minimum height=13pt,yshift=4pt,
 inner sep=1pt,text=white,font=\sffamily\bfseries\fontsize{6.5}{7.5}\selectfont]
 at (E.south) {Success};
\newcommand{\cswatch}[2]{{\setlength{\fboxsep}{0pt}\setlength{\fboxrule}{.4pt}%
  \fcolorbox{#1}{#2}{\rule{0pt}{5.5pt}\hspace{5.5pt}}}}
\node[anchor=north west,draw=black!16,rounded corners=5pt,fill=white,
 inner xsep=3pt,inner ysep=3pt,align=center]
 at (0,-155pt) {%
\begin{minipage}{\dimexpr\csLegendWidth-8pt\relax}
\centering\sffamily\fontsize{7}{8.4}\selectfont
\makebox[\linewidth][c]{%
\cswatch{black!25}{casegrey}\hspace{2pt}observation / recovery state\hspace{5pt}%
\cswatch{blue!45}{caseblue}\hspace{2pt}symbolic plan\hspace{5pt}%
\cswatch{green!45!black}{casegreen}\hspace{2pt}grounded command\hspace{5pt}%
\cswatch{orange!75}{caseorange}\hspace{2pt}placeholder hypothesis\hspace{5pt}%
\cswatch{red!60}{casered}\hspace{2pt}grounding failure}
\end{minipage}};
\end{tikzpicture}
\endgroup
    \caption{Case study on a TextWorld Cooking Hard task with an unresolved grilling device. MemPlan introduces a planner-side placeholder \texttt{unknown\_grill\_1}, uses feedback to avoid treating an unrealized symbolic action as executed, replans, and grounds the resolved device \texttt{bbq\_0} to the executable command \texttt{cook red tuna with BBQ}.}
    \label{fig:case_study}
\end{figure}
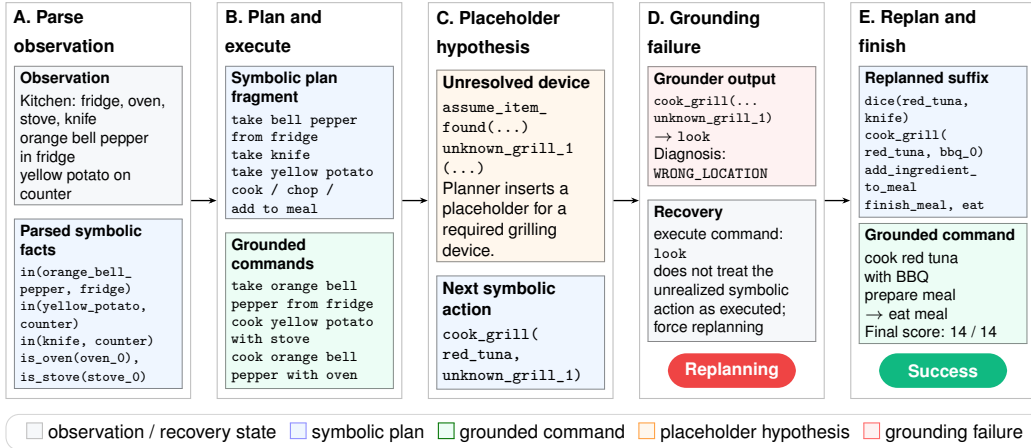

Figure~\ref{fig:case_study} shows a representative TextWorld Cooking Hard run completed by MemPlan in 31 steps with a normalized score of $1.00$. The task requires collecting ingredients, applying preparation and cooking operations, assembling the meal, and eating it. MemPlan first parses textual observations into symbolic memory, then uses Fast Downward to produce a long-horizon PDDL plan.

The critical recovery occurs when the planner needs a grilling device that has not been grounded to a concrete object. MemPlan introduces the placeholder \texttt{unknown\_grill\_1} and plans a \texttt{cook\_grill} action with it. Because the action cannot be grounded to an admissible command at the current step, MemPlan executes an observation command, does not treat the unrealized symbolic action as executed, and replans from updated memory. The replanned suffix resolves the device to \texttt{bbq\_0}, whose executable environment command uses the surface argument \texttt{BBQ}. This allows the agent to execute \texttt{cook red tuna with BBQ} and finish the task.

\section{Limitations}
\label{sec:limitations}

MemPlan has three main limitations. First, it assumes that a task-level PDDL domain is available and covers the relevant object types, predicates, action schemas, preconditions, and effects. Although TextWorld domains can be generated and repaired offline before evaluation accurately, MemPlan does not automate domain construction inside the online agent; incorrect or incomplete domain rules can still lead to invalid symbolic plans. Second, negative memory currently uses hard exclusions: once feedback marks a candidate hypothesis as refuted, the candidate is removed from later PDDL problems. This avoids repeated failed checks, but it can be brittle when feedback is ambiguous or incorrectly parsed, because a valid candidate may be excluded too early. Future work could replace this hard exclusion with confidence-weighted negative memory, where weak or ambiguous feedback lowers a candidate's priority instead of removing it immediately. Third, our evaluation is limited to simulated text environments with benchmark-specific action spaces and, when available, admissible commands. We do not evaluate safety-critical deployment, physical robotics, or open-ended environments without constrained command sets.



\section{Conclusion}
\label{sec:conclusion}

This work shows that a fixed-domain classical planner can be used effectively in partially observable text environments when interaction memory is made part of PDDL problem construction. MemPlan represents unresolved task-relevant information as candidate hypotheses, uses positive graph memory to assign planner-visible action costs, and converts execution-refuted assumptions into exclusions for later planning calls. This design keeps the LLM out of the online planning role and uses it only for constrained parsing and grounding. Across ALFWorld, TextWorld, and Robotouille, the results show that memory-conditioned PDDL construction improves task completion and step efficiency over language-only and PDDL-aware baselines. More broadly, MemPlan suggests a concrete role for memory in neural--symbolic agents: interaction memory can be compiled into planner-visible costs and constraints, rather than only retrieved as language context.

\begin{ack}
\textbf{Funding.} This work was supported by KAUST -- Center of Excellence for Generative AI, under award number 5940.
This work has also received funding from the European Union under the CEREBRA-AI project (Grant Agreement No. 101299046). Views and opinions expressed are, however, those of the author(s) only and do not necessarily reflect those of the European Union or the European Health and Digital Executive Agency (HaDEA). Neither the European Union nor HaDEA can be held responsible for them.

\textbf{Competing interests.} The authors declare no competing interests.
\end{ack}

\printrevisedbibliography

\newpage
\appendix

\section{Experimental Details}
\label{app:experimental_details}

This appendix provides the experimental details for Section~\ref{sec:experiments}. It includes benchmark scope, episode budgets, metric definitions, statistical reporting, PDDL and planner settings, hypothesis-cost construction, and online token accounting.

\subsection{Environment Roles and Evaluation Scope}
\label{app:environment_roles}

We evaluate MemPlan on three interactive text environments. Each environment is used for a different experimental purpose.

\paragraph{ALFWorld.}
ALFWorld is the main end-to-end benchmark. It contains household manipulation tasks that require object search, receptacle reasoning, object-state changes, command grounding, and long-horizon execution. We evaluate on 134 ALFWorld episodes across six task types: clean, cool, examine, heat, put, and puttwo. Results are reported by task type and by macro-average over task types.

\paragraph{TextWorld.}
TextWorld is used to compare MemPlan against memory-augmented LLM agents and to evaluate cross-task use of the learned language--symbol interface. We evaluate on 7 TextWorld task instances, reported as three groups in Figure~\ref{fig:textworld_main}: Treasure Hunt Hard, Cooking Hard, and Cleaning.

\paragraph{Robotouille.}
Robotouille is used to evaluate synchronous and asynchronous long-horizon planning. We evaluate 151 task-seed runs. For each method and setting, success is first averaged over seeds within each task configuration and then macro-averaged across task configurations. Synchronous and asynchronous settings are reported separately. Each compared method uses the same task budget and termination rule for a given Robotouille task.

\subsection{Episode Budgets and Evaluation Units}
\label{app:episode_budgets}

All compared methods use the same episode budget within each environment and task.

\begin{table}[h]
\centering
\small
\caption{Evaluation runs and episode budgets.}
\begin{tabular}{lcc}
\toprule
Environment & Evaluation unit & Episode budget \\
\midrule
TextWorld & 7 task instances & task-dependent limits \\
ALFWorld & 134 episodes & 150 steps \\
Robotouille & 151 completed multi-seed runs & derived from each task's optimal-step budget \\
\bottomrule
\end{tabular}
\label{tab:appendix_episode_budgets}
\end{table}

For ALFWorld, an episode is counted as successful if the environment reports task completion within 150 environment steps. For TextWorld, we follow the task-dependent limits and normalized-score protocol used by the compared memory-agent baselines. For Robotouille, a run is counted as successful if all required subtasks are completed within the task-specific budget.

\begin{table}[h]
\centering
\small
\caption{TextWorld evaluation grouping.}
\begin{tabular}{lll}
\toprule
Group in Figure~\ref{fig:textworld_main} & Instance identifiers & Role in transfer evaluation \\
\midrule
Treasure Hunt Hard & hunt, hunt\_hard & fine-tuning task family \\
Cooking Hard & cook, cook\_hard, cook\_hardest, cook\_rl\_baseline & fine-tuning task family \\
Cleaning & clean\_3x3 & held-out task family \\
\bottomrule
\end{tabular}
\label{tab:textworld_instance_groups}
\end{table}

\subsection{Metric Definitions and Statistical Reporting}
\label{app:statistical_reporting}

\paragraph{Task success.}
Task success is $1$ if the environment reports task completion within the episode budget and $0$ otherwise. For ALFWorld, success is reported by task type and as a macro-average over the six task types. For Robotouille, success is reported separately for synchronous and asynchronous settings as a task-configuration macro-average after averaging over seeds.

\paragraph{Episode length.}
Episode length is the number of executed environment commands before success or budget exhaustion. Observation fallbacks, such as \texttt{look}, are counted as environment steps because they consume interaction budget.

\paragraph{Normalized score.}
For TextWorld, normalized score follows the protocol used by the compared memory-agent baselines. Scores are first computed for each task instance and then aggregated into the task groups reported in Figure~\ref{fig:textworld_main}.

\paragraph{Invalid actions.}
Invalid action rate is the fraction of executed commands rejected by the environment. For MemPlan, if a grounding output does not exactly match an admissible command, command normalization and admissible-command matching are applied before execution. If no admissible match exists and the command is executed and rejected by the environment, it is counted as invalid.

\paragraph{Repeated hypothesis checks.}
Repeated checks count targeted inspections or information-gathering actions that test a candidate already refuted earlier in the same episode. For example, checking the same unresolved object-location hypothesis after a prior ``not found'' feedback increments this metric.

\paragraph{Planner failures.}
Planner failures count cases where Fast Downward returns no plan, reaches the configured planning timeout, or returns a plan whose first symbolic action cannot be matched to any admissible command and cannot be replaced by an admissible observation command.

\paragraph{Online LLM tokens.}
Online LLM tokens count prompt and completion tokens used during test-time interaction. Fine-tuning tokens and offline data-construction tokens are excluded. Symbolic planner calls and deterministic PDDL construction are excluded because they do not invoke an LLM.

For success rates, we report point estimates and bootstrap confidence intervals over evaluation episodes or runs. For ALFWorld, bootstrap resampling is performed over the 134 episodes, with task-type macro-averages recomputed for each bootstrap sample. For Robotouille, bootstrap resampling is performed over completed multi-seed runs separately for synchronous and asynchronous settings. For TextWorld, because only 7 task instances are used, we report per-instance scores and group averages.

Unless otherwise stated, confidence intervals are two-sided $95\%$ bootstrap intervals computed with 10{,}000 bootstrap resamples. When two methods are evaluated on matched episodes, we additionally report paired differences for success and episode length. These intervals quantify variation over the fixed benchmark split.

\begin{table}[h]
\centering
\small
\caption{Statistical reporting summary.}
\resizebox{\linewidth}{!}{
\begin{tabular}{lcccc}
\toprule
Environment & Resampling unit & Main metric & Main-text estimate & Appendix statistics \\
\midrule
ALFWorld & episode & success, length, tokens & yes & 95\% CI, paired differences \\
TextWorld & task instance & normalized score & yes & per-instance scores, group averages \\
Robotouille & completed run & success & yes & 95\% CI by sync/async setting \\
Ablations & episode or run & success, length, diagnostics & yes & 95\% CI, paired differences \\
\bottomrule
\end{tabular}
}
\label{tab:statistical_reporting_summary}
\end{table}

\subsection{PDDL Domains and Planner}
\label{app:pddl_planner}

MemPlan separates fixed PDDL domains from online PDDL problem construction. The domain specifies object types, predicates, action schemas, preconditions, and effects. The problem file specifies the current objects, initial facts, goal facts, memory-conditioned candidate hypotheses, action costs, and exclusion constraints. During evaluation, MemPlan reconstructs only the problem file online; the domain file remains fixed.

The source of the fixed domain differs by benchmark. For ALFWorld, we use the benchmark-provided ALFRED PDDL domain released with ALFWorld. For Robotouille, we construct PDDL domains from public domain metadata and transition specifications. For TextWorld, because the benchmark does not provide PDDL domains for the evaluated task families, we generate domains offline before evaluation. The domain-generation prompt contains environment-level task rules and action descriptions only; it does not include test-instance maps, object placements, oracle trajectories, interaction logs, or evaluation feedback. GPT-5.1 drafts each domain, and validator or planner error messages are used for one offline repair round when needed. In our protocol, $5/7$ TextWorld domains are valid on the first attempt, and the remaining two are repaired successfully after one feedback round. The resulting domains are fixed during evaluation and are not updated online.

MemPlan uses Fast Downward~\citep{helmert2006fast} for symbolic planning. Generated problem files include:
\begin{verbatim}
(:metric minimize (total-cost))
\end{verbatim}

The planner configuration is satisficing, so action costs bias hypothesis selection without guaranteeing globally optimal-cost plans. After each planner call, MemPlan executes only the first grounded command, observes textual feedback, updates memory, and replans when needed.

When an observation fallback is used, the unrealized symbolic action is not treated as executed. Memory is updated only from environment feedback.

\subsection{Example Compiled PDDL Problem}
\label{app:compiled_problem_example}

The following listing shows one complete problem file generated by \textsc{BuildProblem} for the TextWorld Treasure Hunt run \texttt{pddl\_output/hunt/job\_46727745/problem\_step\_5.pddl}. The task is to obtain the target object \texttt{o\_2}. This time step is useful because the problem contains planner-side placeholders, observation-derived memory facts, planner-visible candidate costs, and negative-memory exclusions from already searched rooms.

Negative memory is stored internally as typed triples over unresolved placeholders and candidate values. During PDDL problem construction, these triples are compiled into the predicates required by the fixed domain. In this Treasure Hunt domain, a refuted room candidate is represented by the unary predicate \texttt{searched($r$)} because the exclusion only needs to block future checks of room $r$. Thus an internal record such as \texttt{(unknown\_o\_2, searched, r\_0)} is compiled as \texttt{(searched r\_0)} in the problem file.

{\scriptsize
\begin{verbatim}
(define (problem step_5)
  (:domain textworld)
  (:objects
    P - p
    I - i
    r_0 r_1 r_3 r_2 r_6 r_8 r_10 r_9 r_11 r_4 r_5 r_7
    unknown_r_1 loc_limbo - r      ; placeholder room + limbo for unknown locations
    c_0 c_1 c_10 c_11 c_2 c_3 c_4 c_5 c_6 c_7 c_8 c_9 - c
    k_0 k_1 k_2 k_3 o_2 unknown_k_1 - k
    o_0 o_1 unknown_o_1 unknown_o_2 - o
    unknown_f_1 unknown_f_2 - f
    unknown_d_1 - d
    unknown_t_1 unknown_t_2 - t
  )
  (:init
    (= (total-cost) 0)

    ; Positive memory: validated observation facts become ordinary init facts.
    (at P r_5)
    (at c_5 r_5)
    (at c_7 r_7)
    (at c_4 r_4)
    (at c_3 r_3)
    (at c_10 r_10)
    (at c_1 r_1)
    (at c_9 r_9)
    (at c_8 r_8)
    (at c_11 r_11)
    (at c_0 r_0)
    (at c_6 r_6)
    (at c_2 r_2)
    (locked c_2)
    (locked c_11)
    (locked c_5)
    (locked c_9)
    (locked c_0)
    (locked c_8)
    (locked c_7)
    (locked c_3)
    (locked c_1)
    (locked c_6)
    (locked c_10)
    (locked c_4)
    (in o_2 c_0)
    (in o_1 c_2)
    (in k_3 c_9)
    (in k_1 c_6)
    (in o_0 c_6)
    (in k_0 I)
    (in k_2 c_2)
    (match k_3 c_0)
    (match k_2 c_9)
    (match k_1 c_2)
    (match k_0 c_6)
    (match o_2 unknown_d_1)
    (is_target_object o_2)
    (is_player_inventory I)

    ; Placeholders for unresolved objects remain in limbo until assumed or grounded.
    (at unknown_o_2 loc_limbo)
    (at k_2 loc_limbo)
    (at unknown_f_1 loc_limbo)
    (at k_3 loc_limbo)
    (at unknown_t_1 loc_limbo)
    (at unknown_f_2 loc_limbo)
    (at o_2 loc_limbo)
    (at k_1 loc_limbo)
    (at unknown_k_1 loc_limbo)
    (at o_0 loc_limbo)
    (at unknown_o_1 loc_limbo)
    (at unknown_d_1 loc_limbo)
    (at k_0 loc_limbo)
    (at unknown_r_1 loc_limbo)
    (at unknown_t_2 loc_limbo)
    (at o_1 loc_limbo)
    (is_unknown_area loc_limbo)
    (is_frontier unknown_f_1)

    ; Positive memory supplies planner-visible candidate costs.
    (= (hint_cost r_10) 1)
    (= (hint_cost loc_limbo) 1)
    (= (hint_cost r_9) 1)
    (= (hint_cost r_8) 1)
    (= (hint_cost r_2) 1)
    (= (hint_cost r_5) 1)
    (= (hint_cost r_11) 1)
    (= (hint_cost r_3) 1)
    (= (hint_cost unknown_r_1) 1)
    (= (hint_cost r_7) 1)
    (= (hint_cost r_6) 1)
    (= (hint_cost r_4) 1)
    (= (hint_cost r_1) 1)
    (= (hint_cost r_0) 1)

    ; Negative memory: internal refutation triples are compiled into domain-level exclusion predicates.
    ; Here, searched(r) blocks repeated checks of a refuted room candidate.
    (searched r_0)
    (searched r_5)
    (searched r_1)

    ; Navigation graph recovered from interaction memory.
    (free r_3 r_6)
    (west_of r_2 r_3)
    (east_of r_8 r_10)
    (free r_3 r_2)
    (free r_7 r_6)
    (free r_1 r_5)
    (free r_6 r_11)
    (free r_6 r_3)
    (free r_9 r_8)
    (free r_5 r_1)
    (north_of r_1 r_2)
    (free r_6 r_7)
    (west_of r_1 r_0)
    (free r_11 r_6)
    (free r_2 r_3)
    (east_of r_3 r_2)
    (free r_2 r_4)
    (free r_4 r_2)
    (west_of r_3 r_6)
    (free r_10 r_8)
    (east_of r_6 r_3)
    (free r_8 r_9)
    (free r_7 r_8)
    (free r_8 r_7)
    (south_of r_11 r_6)
    (south_of r_4 r_2)
    (south_of r_2 r_1)
    (west_of r_10 r_8)
    (north_of r_5 r_1)
    (south_of r_7 r_8)
    (free r_8 r_10)
    (north_of r_7 r_6)
    (north_of r_6 r_11)
    (east_of r_9 r_8)
    (east_of r_0 r_1)
    (free r_1 r_0)
    (south_of r_1 r_5)
    (free r_1 r_2)
    (north_of r_8 r_7)
    (south_of r_6 r_7)
    (free r_0 r_1)
    (free r_2 r_1)
    (west_of r_8 r_9)
    (north_of r_2 r_4)
  )
  (:goal (and (in o_2 I)))
  (:metric minimize (total-cost))
)
\end{verbatim}
}

The corresponding fixed domain action consumes the cost and exclusion facts supplied by the problem file:

{\scriptsize
\begin{verbatim}
(:action assume_item_found
  :parameters (?P - P ?o - o ?r - r ?unk - r)
  :precondition (and
    (at ?P ?r)
    (at ?o ?unk)
    (is_unknown_area ?unk)
    (not (= ?r ?unk))
    (not (searched ?r)))       ; compiled negative memory blocks refuted room candidates
  :effect (and
    (increase (total-cost) (hint_cost ?r)) ; graph cost biases hypotheses
    (not (at ?o ?unk))
    (at ?o ?r)
    (searched ?r)))     ; record that this room candidate has been checked
\end{verbatim}
}

\subsection{Graph-Memory Hypothesis Cost Construction}
\label{app:hypothesis_cost_details}

For each unresolved placeholder $z$, MemPlan constructs candidate hypotheses $h=(z,r,v)$ from entities and relations observed in the current episode. Candidate values are drawn from observed locations, containers, receptacles, objects, states, and task-related entities in positive memory. Candidates already refuted by negative memory are removed before PDDL problem construction.
The internal negative-memory records are triples for consistency with hypothesis representation, but \textsc{BuildProblem} compiles them into the arity required by the target PDDL domain. For example, a placeholder-specific refutation may compile to \texttt{not\_at($z$,$v$)}, while a domain that only needs to block repeated room checks may compile the same evidence to a unary predicate such as \texttt{searched($v$)}.

For each remaining candidate $h=(z,r,v)$, MemPlan computes a relevance score $\mathrm{Score}_t(h)\in[0,1]$ as a text-based graph-memory retrieval score. The query string is
\[
q_t(z)=\texttt{Goal: }g\ \texttt{; Missing: }z .
\]
The candidate document is
\[
d_t(h)=\texttt{Candidate: }(z,r,v)\ \texttt{; Context: }\mathrm{Linearize}(N_1(v;M_t^+)),
\]
where $N_1(v;M_t^+)$ is the one-hop neighborhood of $v$ in the positive-memory graph. The function $\mathrm{Linearize}(\cdot)$ serializes neighboring triples in a deterministic order. We encode $q_t(z)$ and $d_t(h)$ with the frozen mContriever encoder (checkpoint \texttt{facebook/mcontriever})~\rev{\citep{izacard2022contriever}} and compute cosine similarity:
\[
\mathrm{sim}_t(h)=
\cos\left(
\mathrm{Enc}(q_t(z)),
\mathrm{Enc}(d_t(h))
\right).
\]
The score is normalized over the current candidate set:
\[
\mathrm{Score}_t(h)=
\frac{\mathrm{sim}_t(h)-\min_{h'\in\mathcal{H}_t}\mathrm{sim}_t(h')}
{\max_{h'\in\mathcal{H}_t}\mathrm{sim}_t(h')-\min_{h'\in\mathcal{H}_t}\mathrm{sim}_t(h')+\epsilon}.
\]
We set $\epsilon=10^{-8}$ for numerical stability. The score is used only to order hypothesis-checking actions. It does not generate symbolic plans, add memory facts, or decide whether a candidate is true.

MemPlan converts the score into a PDDL action cost:
\[
\kappa_t(h)
=
\kappa_{\mathrm{base}}
+
\left\lceil
\lambda \cdot \max(0,\tau-\mathrm{Score}_t(h))
\right\rceil ,
\]
where $\kappa_{\mathrm{base}}$ is the base action cost, $\tau$ is the relevance threshold, and $\lambda$ controls the penalty scale. Physical actions use $\kappa_{\mathrm{base}}$. Hypothesis-checking actions receive larger costs when their relevance scores are lower than $\tau$.

The generated PDDL problem includes:
\begin{verbatim}
(:metric minimize (total-cost))
\end{verbatim}
Thus, the planner is biased toward checking hypotheses with stronger positive-memory support. Since the planner configuration is satisficing, these costs guide search without guaranteeing a globally minimum-cost plan. All cost parameters are fixed before test-time evaluation and are not tuned on test episodes. Negative memory can still remove a high-scoring candidate after it is refuted by execution feedback.

\subsection{\rev{Positive-Memory Hyperparameter Sensitivity}}
\label{app:sensitivity}
\rev{We tested all nine combinations of $\tau\in\{0.40,0.50,0.60\}$ and $\lambda\in\{5,10,20\}$, with $\kappa_{\mathrm{base}}=1$. Across the evaluated TextWorld tasks, all nine settings achieved a normalized score of 1.00; mean episode lengths ranged from 31.0 to 34.1 steps. These results support the role of positive memory in ordering hypothesis checks and improving interaction efficiency. The ability to complete the tasks also depends on placeholders, negative memory, and the fixed PDDL domain.}

\subsection{Online Token Accounting}
\label{app:token_accounting}

Online LLM tokens count prompt and completion tokens used during test-time interaction. For language-only baselines, this includes action-generation calls. For ReAct-style baselines, this includes reasoning and action-generation text. For LLM-DP, this includes online belief-state generation and action-selection text required during test-time interaction. For MemPlan, online LLM tokens come from Qwen2.5-7B interface calls for observation parsing and command grounding. Symbolic planner calls and deterministic PDDL construction are excluded because they do not invoke an LLM.

Offline costs are reported separately from online interaction costs. This includes supervised fine-tuning of the language--symbol interface and offline TextWorld domain generation/repair. The online token tables, therefore, measure test-time interaction cost, not one-time preprocessing cost.

Supervised fine-tuning cost is reported separately from online test-time token usage.

\begin{table}[h]
\centering
\small
\caption{Online token accounting. MemPlan uses language-model calls only for parsing and grounding.}
\begin{tabular}{lccc}
\toprule
Model & Action-generation tokens & Interface tokens & Planner tokens \\
\midrule
GPT-4o & yes & no & no \\
GPT-4o + Domain & yes & no & no \\
ReAct & yes & no & no \\
LLM-DP & yes & no & no \\
MemPlan & no & yes & no \\
\bottomrule
\end{tabular}
\label{tab:token_accounting}
\end{table}

\begin{table}[h]
\centering
\small
\caption{Per-episode online efficiency statistics.}
\resizebox{\linewidth}{!}{
\begin{tabular}{lcccc}
\toprule
Model & Steps / episode & LLM calls / episode & Tokens / episode & 95\% CI for tokens / episode \\
\midrule
ReAct (GPT-4o) & 38.18 & 39.78 & 37.9k & [34.8k, 41.2k] \\
LLM-DP & 13.16 & 13.16 & 4.7k & [4.2k, 5.3k] \\
MemPlan & 6.91 & 16.87 & 6.9k & [6.4k, 7.5k] \\
\bottomrule
\end{tabular}
}
\label{tab:appendix_efficiency_ci}
\end{table}

\section{Interface Training and Validation Details}
\label{app:interface_training}

\subsection{Unified Interface Model}
\label{app:unified_interface}

MemPlan uses Qwen2.5-7B fine-tuned as a unified schema-conditioned language--symbol interface. The same frozen model handles two task formats: observation-to-fact parsing and symbolic-action-to-command grounding.

\paragraph{Observation-to-fact parsing.}
The input contains a textual observation, the domain predicate schema, and an object glossary. The output is a set of PDDL-compatible facts. Outputs are checked against valid predicate names and typed object identifiers before they are used to update memory or construct the next PDDL problem.

\paragraph{Symbolic-action-to-command grounding.}
The input contains a symbolic planner action, the current textual observation, and the admissible command set. The output is an executable text command. This turns command generation into candidate selection from admissible commands.

\subsection{Training Data}
\label{app:interface_training_data}

The Qwen2.5-7B interface is fine-tuned on 17{,}000 observation--fact examples and 9{,}600 symbolic-action--command examples. Observation--fact labels come from simulator annotations. Symbolic-action--command labels are constructed through heuristic alignment between planner actions, oracle trajectories, and admissible commands.

\begin{table}[h]
\centering
\small
\caption{Training data for the unified Qwen2.5-7B language--symbol interface.}
\resizebox{\linewidth}{!}{
\begin{tabular}{lcc}
\toprule
Training task format & Number of examples & Label source \\
\midrule
Observation-to-fact parsing & 17{,}000 & simulator annotations \\
Symbolic-action-to-command grounding & 9{,}600 & planner-action / oracle-trajectory / admissible-command alignment \\
\bottomrule
\end{tabular}
}
\label{tab:interface_training_data}
\end{table}

\subsection{TextWorld Interface Transfer Setting}
\label{app:textworld_interface_transfer}

For TextWorld, the Qwen2.5-7B interface is fine-tuned only on cook and hunt task families. After fine-tuning, the interface is frozen and used inside the full MemPlan loop for the remaining TextWorld task types. No examples from target task types are used for supervised fine-tuning.

\begin{table}[h]
\centering
\small
\caption{TextWorld interface-transfer setting. Cooking Hard and Treasure Hunt Hard correspond to the fine-tuning task families, while Cleaning evaluates cross-task use of the frozen interface.}
\resizebox{\linewidth}{!}{%
\begin{tabular}{lll}
\toprule
Fine-tuning task families & Evaluation task groups in Figure~\ref{fig:textworld_main} & Interface update during evaluation \\
\midrule
cook, hunt & Cooking Hard, Treasure Hunt Hard, Cleaning & none \\
\bottomrule
\end{tabular}%
}
\label{tab:textworld_transfer_setting}
\end{table}

\subsection{Interface Validation Splits}
\label{app:interface_validation_splits}

The interface is trained only on TextWorld cook and hunt task families. The ALFWorld and Robotouille validation splits reported in Table~\ref{tab:interface_validation_metrics} are used only for post-training diagnostic evaluation of parsing and grounding behavior. They are not used for supervised fine-tuning, checkpoint selection, prompt tuning, threshold tuning, or any test-time adaptation. The same frozen checkpoint is used in all end-to-end evaluations.

\subsection{Output Validation}
\label{app:interface_output_validation}

For observation parsing, predicted facts are accepted only if they use valid predicates and typed object identifiers from the current domain schema and object glossary. Invalid facts are discarded before memory update.

For command grounding, the model selects an admissible command. If the output does not exactly match an admissible command, we apply command normalization and admissible-command matching. Commands rejected by the environment are counted in the invalid-action metric.

\subsection{Interface Validation Metrics}
\label{app:interface_validation_metrics}

We evaluate the interface separately from the end-to-end planning loop. The interface is trained only on TextWorld cook and hunt task families; ALFWorld and Robotouille validation examples are used only for post-training diagnostic evaluation and are not used for supervised fine-tuning, checkpoint selection, prompt tuning, threshold tuning, or test-time adaptation. Observation parsing is evaluated using precision, recall, and F1 over accepted PDDL-compatible facts. Command grounding is evaluated using top-1 admissible-command accuracy. We also report schema rejection and environment rejection rates to measure invalid parsed facts and commands rejected by the environment.

\begin{table}[h]
\centering
\small
\caption{Validation metrics for the schema-conditioned language--symbol interface.}
\resizebox{\linewidth}{!}{
\begin{tabular}{lcccccc}
\toprule
Environment / split & Parsing precision & Parsing recall & Parsing F1 & Grounding top-1 & Schema rejection & Env. rejection \\
\midrule
ALFWorld validation & 0.94 & 0.92 & 0.93 & 0.96 & 0.03 & 0.01 \\
TextWorld validation & 0.96 & 0.94 & 0.95 & 0.97 & 0.02 & 0.01 \\
Robotouille validation & 0.91 & 0.89 & 0.90 & 0.94 & 0.04 & 0.02 \\
\bottomrule
\end{tabular}
}
\label{tab:interface_validation_metrics}
\end{table}

\subsection{Interface Error Categories}
\label{app:interface_error_categories}

We classify interface errors into the following categories.

\paragraph{Parser omission.}
A true observation-supported fact is missing from the parser output.

\paragraph{Parser hallucination.}
The parser outputs a fact that is not supported by the observation. Schema validation removes malformed or type-invalid hallucinations before they can update memory.

\paragraph{Entity mismatch.}
The parser or grounder selects the wrong object identifier, receptacle, or location among valid candidates.

\paragraph{Predicate mismatch.}
The parser uses a domain-valid predicate whose relation type does not match the observation.

\paragraph{Grounding mismatch.}
The grounder selects an admissible command whose surface form is executable but does not correspond to the intended symbolic action.

\paragraph{No safe grounding.}
The symbolic action contains an unresolved placeholder or context-dependent object that cannot be grounded to an admissible command. In this case, MemPlan executes an admissible observation command when available. The unrealized symbolic action is not treated as executed, and memory is updated only from environment feedback.

\begin{table}[h]
\centering
\small
\caption{Interface error breakdown.}
\resizebox{\linewidth}{!}{
\begin{tabular}{lcccccc}
\toprule
Split & Parser omission & Parser hallucination & Entity mismatch & Predicate mismatch & Grounding mismatch & No safe grounding \\
\midrule
ALFWorld & 28 & 14 & 18 & 10 & 20 & 10 \\
TextWorld & 24 & 12 & 16 & 9 & 28 & 11 \\
Robotouille & 30 & 16 & 20 & 12 & 14 & 8 \\
\bottomrule
\end{tabular}
}
\label{tab:interface_error_breakdown}
\end{table}

\subsection{Training Hyperparameters}
\label{app:training_hyperparameters}

Table~\ref{tab:interface_hyperparameters} reports interface training hyperparameters.

\begin{table}[h]
\centering
\small
\caption{Training hyperparameters for the unified Qwen2.5-7B interface model.}
\begin{tabular}{lp{0.62\linewidth}}
\toprule
Item & Value \\
\midrule
Base model & Qwen2.5-7B-Instruct \\
Fine-tuning method & LoRA, rank 16, alpha 32, dropout 0.05 \\
Training epochs & 2 \\
Learning rate & $5\times 10^{-5}$ \\
Batch size & 4 per device, gradient accumulation 4 \\
Maximum input length & 131{,}072 tokens tokenizer limit \\
Maximum output length & no separate cap; target shares the 131{,}072-token sequence limit \\
Optimizer & AdamW, linear schedule, no warmup, weight decay 0 \\
Checkpoint selection & final checkpoint after 2 epochs; intermediate checkpoints saved every 166 steps \\
\bottomrule
\end{tabular}
\label{tab:interface_hyperparameters}
\end{table}

\section{Baseline Details}
\label{app:baseline_details}

\subsection{Baseline Configurations}
\label{app:baseline_configurations}

We compare against language-only baselines, including GPT-4o, GPT-4o with domain descriptions, and ReAct~\citep{yao2023react}; PDDL-aware baselines, including LLM-DP~\citep{dagan2023dynamic}; and TextWorld memory-agent baselines from AriGraph~\citep{anokhin2025arigraph}. Baseline prompts, model versions, admissible-command access, domain-information access, history windows, and stopping rules are fixed before evaluation.

\subsection{PDDLEGO and PDDLEGO+}
\label{app:pddlego_protocol}

PDDLEGO and PDDLEGO+ are the closest prior methods that iteratively construct or refine PDDL representations during interaction. \rev{The published PDDLEGO+ results use a different trial set and manually implemented fine-grained simulation errors for online PDDL repair.} \rev{We evaluate PDDLEGO+ on the 134 ALFWorld episodes in two configurations (Table~\ref{tab:alfworld_comparison}). PDDLEGO+-Native retains the original fine-grained repair feedback. PDDLEGO+-Shared receives the same standard textual environment feedback available to MemPlan, without the additional error channel. Apart from feedback access, both use the task goals, episode budget, admissible-command access, and stopping criteria in our evaluation protocol. Both retain iterative PDDL formalization, planning, execution, and refinement. We account for all online PDDL generation and repair calls in token usage.}

\subsection{Common Evaluation Protocol}
\label{app:baseline_common_protocol}

All baselines are evaluated with the same task goals, textual observations, episode limits, and admissible-command access when the benchmark exposes admissible commands. Baselines differ in how they use language models, memory, domain information, and symbolic planning.

The stopping rule is shared across methods: the episode terminates when the environment reports success, the method reaches the episode budget, or the benchmark-specific failure condition is triggered. Prompt templates, model versions, decoding parameters, history-window sizes, and domain-description inputs are fixed before evaluation and are not tuned on test episodes.

\begin{table}[h]
\centering
\small
\caption{Information available to each method at test time. ``Benchmark-dependent'' means that admissible-command access is used only when the corresponding benchmark exposes it and is applied consistently across compared methods within that benchmark.}
\resizebox{\linewidth}{!}{
\begin{tabular}{lccccc}
\toprule
Model & Goal / observation & Interaction history & Admissible commands & Domain information & Symbolic planner \\
\midrule
GPT-4o & yes & yes & benchmark-dependent & no & no \\
GPT-4o + Domain & yes & yes & benchmark-dependent & natural-language domain description & no \\
ReAct & yes & recent history & benchmark-dependent & no & no \\
LLM-DP & yes & yes & benchmark-dependent & PDDL-aware belief planning & yes \\
PDDLEGO / PDDLEGO+ & yes & yes & benchmark-dependent & iterative PDDL construction/refinement & yes \\
MemPlan & yes & symbolic memory & yes when exposed & fixed PDDL domain and predicate schema & yes \\
\bottomrule
\end{tabular}
}
\label{tab:baseline_inputs}
\end{table}

These comparisons should be interpreted as system-level comparisons among language-only, PDDL-aware, and memory-conditioned PDDL-planning agents. MemPlan assumes a fixed PDDL domain and studies how interaction memory can be compiled into online PDDL problem construction.

\subsection{Language-Only Baselines}
\label{app:language_only_baselines}

\paragraph{GPT-4o.}
GPT-4o receives the task goal, current observation, and interaction history. It outputs one text command per step. When admissible commands are provided by the benchmark, they are included in the prompt using the same admissible-command access policy used by other language-only baselines.

\paragraph{GPT-4o + Domain.}
GPT-4o + Domain receives the same input as GPT-4o, plus a natural-language description of task-relevant action types, object types, and environment rules. It does not receive a machine-readable PDDL problem file or planner output.

\paragraph{ReAct.}
ReAct~\citep{yao2023react} interleaves reasoning and action generation. The model receives the task goal, current observation, and recent interaction history, then outputs one command to execute. The reasoning text is counted in online LLM tokens. \rev{ReAct (GPT-5.1) follows the same ALFWorld and Robotouille evaluation protocols, including the same task configurations and seeds where applicable (Tables~\ref{tab:alfworld_comparison} and~\ref{tab:robotouille_main}).}

\subsection{PDDL-Aware Baselines}
\label{app:pddl_aware_baselines}

\paragraph{LLM-DP.}
LLM-DP~\citep{dagan2023dynamic} uses LLM-generated belief states for symbolic planning. We evaluate it with the same episode limits and environment observations used for MemPlan. Its online LLM tokens include belief-state generation and action-selection text required during test-time interaction.

\paragraph{PDDLEGO and PDDLEGO+.}
PDDLEGO~\citep{zhang2024pddlego} and PDDLEGO+~\citep{gong2025pddlego} iteratively construct or refine PDDL representations in partially observable textual environments. \rev{We report the same-protocol PDDLEGO+-Native and PDDLEGO+-Shared comparisons in Table~\ref{tab:alfworld_comparison}; they differ in whether the original fine-grained repair errors or standard environment feedback is used.}

\subsection{TextWorld Memory Baselines}
\label{app:textworld_memory_baselines}

For TextWorld, we compare MemPlan with memory-augmented LLM agents when published protocols are compatible. These baselines include full-history context, summary memory, retrieval memory, Reflexion~\citep{shinn2023reflexion}, Simulacra-style simulation memory, and AriGraph~\citep{anokhin2025arigraph}. These baselines use memory to support LLM decision-making. MemPlan uses memory differently: it compiles graph-memory relevance into PDDL hypothesis-action costs and compiles execution-refuted candidates into exclusions for later problem construction.

\subsection{Prompt and Decoding Configuration}
\label{app:prompt_decoding_config}

Table~\ref{tab:prompt_decoding_config} summarizes the prompt and decoding configuration used for LLM baselines and for the MemPlan interface. The exact prompt templates are included in the released code.

\begin{table}[h]
\centering
\small
\caption{Prompt and decoding configuration.}
\resizebox{\linewidth}{!}{
\begin{tabular}{lcccc}
\toprule
Model / component & Model version & Temperature & Max output tokens & History window \\
\midrule
GPT-4o & gpt-4o & 0.0 & 512 & full episode history \\
GPT-4o + Domain & gpt-4o & 0.0 & 512 & full episode history + domain description \\
ReAct (GPT-4o) & gpt-4o & 0.0 & 512 & full episode history \\
ReAct (LLM-DP) & gpt-4o & 0.0 & 512 & belief-state history \\
MemPlan parser & Qwen2.5-7B fine-tuned & 0.0 & 1024 & current observation \\
MemPlan grounder & Qwen2.5-7B fine-tuned & 0.0 & 1024 & current observation + admissible commands \\
\bottomrule
\end{tabular}
}
\label{tab:prompt_decoding_config}
\end{table}

\section{Ablation Details}
\label{app:ablation_details}

Ablations are evaluated on the same seven TextWorld task instances used in Figure~\ref{fig:textworld_main}: \texttt{hunt}, \texttt{hunt\_hard}, \texttt{cook}, \texttt{cook\_hard}, \texttt{cook\_hardest}, \texttt{cook\_rl\_baseline}, and \texttt{clean\_3x3}. Each variant is run with three seeds per instance, yielding 21 runs per variant and 147 runs in total. All variants use the same PDDL domains, planner settings, candidate-generation rules, and interaction budgets unless the ablated component changes them.

\subsection{Ablation Variables}
\label{app:ablation_definitions}

\stitle{Uniform-cost hypotheses.}
Replaces graph-memory hypothesis costs with equal costs. Candidate generation, placeholders, negative memory, parsing, grounding, and replanning are unchanged.

\stitle{w/o negative memory.}
Removes searched and refuted triples from PDDL problem construction, allowing previously contradicted hypotheses to be checked again.

\stitle{w/o placeholders.}
Constructs PDDL problems only with observed concrete objects. Required but unobserved entities cannot be represented as planner-internal placeholders.

\stitle{w/o observation parser.}
Removes the fine-tuned observation-to-fact parser while keeping the learned action grounder and all memory/planning components fixed.

\stitle{w/o action grounder.}
Removes the fine-tuned symbolic-action-to-command grounder while keeping the learned observation parser and all memory/planning components fixed.

\stitle{w/o learned interface.}
Removes both fine-tuned interface components.

\subsection{Ablation Metrics}
\label{app:ablation_metrics}

\stitle{Normalized score.}
TextWorld scores are normalized by the maximum achievable score of each task instance and averaged over 21 runs per variant.

\stitle{Steps.}
Number of executed environment commands. Runs that reach the interaction budget without completion are assigned the budgeted step count.

\stitle{Repeated checks.}
Mean number of targeted checks per run that revisit hypotheses already refuted earlier in the same run.

\stitle{Planner dead ends.}
Mean number of cases per run where the constructed PDDL problem yields no usable next symbolic action.

\stitle{Invalid actions.}
Mean number of executed commands per run that are rejected by the environment.

\section{Case-Study Grounding Details}
\label{app:case_study_details}

The case study in Figure~\ref{fig:case_study} contains both planner-side symbols and environment-side command arguments. The placeholder \texttt{unknown\_grill\_1} denotes an unresolved planner-side grilling device. After replanning, the unresolved device is resolved to a concrete planner-side object \texttt{bbq\_0}. The executable environment command uses the surface argument \texttt{BBQ}, yielding the command \texttt{cook red tuna with BBQ}.

This distinction matters because the internal planner object name and the text environment's command argument need not be string-identical. The grounding module maps the symbolic action and its arguments to one command from the admissible command set.

\section{Additional Reproducibility Details}
\label{app:reproducibility}

\paragraph{Compute resources.}
All MemPlan experiments were run on Intel Xeon Platinum 8260 CPU nodes with NVIDIA Tesla V100-SXM2 GPUs. Each V100 has 32~GB GPU memory. The completed Qwen2.5-7B interface fine-tuning job used 8 V100 GPUs, 8 CPU cores, and 32~GB allocated system memory for 62.4 hours, corresponding to approximately 500 GPU-hours. The evaluation jobs used V100-class nodes with 120--160~GB allocated system memory. End-to-end evaluation over the reported ALFWorld, TextWorld, and Robotouille runs took approximately 21 wall-clock hours across the final evaluation campaigns. Fast Downward planning was run on CPU, while language--symbol interface calls used the fine-tuned Qwen2.5-7B model. The reported compute excludes preliminary experiments that did not affect the final reported results.

\paragraph{Code and data availability.}
\rev{We release code, configuration files, PDDL construction scripts, evaluation scripts, and reproduction instructions at \url{https://github.com/tewiSong/MemPlan}.} The experiments use existing benchmark environments: TextWorld, ALFWorld, and Robotouille. The released code documents environment setup, dependency installation, model-interface usage, and commands for reproducing the main tables.

\paragraph{Licenses and existing assets.}
We use existing benchmark environments, models, and software tools, including TextWorld, ALFWorld, Robotouille, Fast Downward, Qwen2.5-7B, GPT-4o, ReAct, and LLM-DP. The released MemPlan code is distributed under the MIT License. TextWorld is released under the MIT License. ALFWorld is released under the MIT License. Robotouille is released under the MIT License. Fast Downward is released under the GNU General Public License v3.0. Qwen2.5-7B-Instruct is released under the Apache License 2.0. We did not vendor external ReAct, LLM-DP, PDDLEGO, or PDDLEGO+ source code in this release; the corresponding baselines are either locally implemented from the published method descriptions or reported only where published protocols are compatible. We cite the corresponding papers in the main text and follow the licenses and terms of use of the original assets.

\paragraph{LLM usage.}
LLMs are used as components of the method and baselines. In MemPlan, the fine-tuned Qwen2.5-7B interface is used for observation-to-fact parsing and symbolic-action-to-command grounding. GPT-4o and ReAct-style baselines use LLMs for online action generation. LLM-DP uses LLM-generated belief states for symbolic planning. No LLM is used to generate or repair MemPlan's PDDL domain during evaluation.

\paragraph{Broader impacts.}
This work studies memory-conditioned symbolic planning for language agents in simulated text environments. A potential positive impact is improved reliability and token efficiency for long-horizon language agents by limiting LLM use to constrained language--symbol interfaces and using explicit symbolic planning. Potential negative impacts include over-reliance on generated symbolic states, failures caused by incorrect parsing or grounding, and misuse of autonomous planning systems in settings where incorrect actions may have real-world consequences. We mitigate these risks by evaluating only in simulated environments, using explicit PDDL constraints, maintaining execution-refuted negative memory, and analyzing failure modes. The method is not evaluated for safety-critical deployment, physical robotics deployment, or decision-making about people.

\newpage
\section*{NeurIPS Paper Checklist}

\begin{enumerate}

\item {\bf Claims}
    \item[] Question: Do the main claims made in the abstract and introduction accurately reflect the paper's contributions and scope?
    \item[] Answer: \answerYes{}
    \item[] Justification: The abstract and introduction state the main contributions: memory-conditioned PDDL planning, positive graph memory for hypothesis costs, negative memory for exclusion constraints, and schema-conditioned language--symbol interfaces. The experimental section evaluates these claims on TextWorld, ALFWorld, and Robotouille in Section~\ref{sec:experiments}.

\item {\bf Limitations}
    \item[] Question: Does the paper discuss the limitations of the work performed by the authors?
    \item[] Answer: \answerYes{}
    \item[] Justification: Appendix~\ref{app:reproducibility} discusses limitations and risks, including failures under partial observability, incorrect parsing or grounding, over-reliance on generated symbolic states, and the fact that the method is not evaluated for safety-critical deployment. 

\item {\bf Theory assumptions and proofs}
    \item[] Question: For each theoretical result, does the paper provide the full set of assumptions and a complete (and correct) proof?
    \item[] Answer: \answerNA{}
    \item[] Justification: The paper does not present theoretical results requiring formal assumptions or proofs. The equations in Section~3 define the planning framework, interface modules, and memory update procedure.

\item {\bf Experimental result reproducibility}
    \item[] Question: Does the paper fully disclose all the information needed to reproduce the main experimental results of the paper to the extent that it affects the main claims and/or conclusions of the paper (regardless of whether the code and data are provided or not)?
    \item[] Answer: \answerYes{}
    \item[] Justification: Section~\ref{sec:experiments} describes the environments, evaluation protocol, baselines, metrics, planner, and interface model. Appendices~\ref{app:experimental_details}, \ref{app:interface_training}, \ref{app:baseline_details}, and \ref{app:reproducibility} provide additional evaluation, training, baseline, compute, and reproduction details, and the anonymized code and data are available at \url{https://github.com/tewiSong/MemPlan}.

\item {\bf Open access to data and code}
    \item[] Question: Does the paper provide open access to the data and code, with sufficient instructions to faithfully reproduce the main experimental results, as described in supplemental material?
    \item[] Answer: \answerYes{}
    \item[] Justification: The paper provides an anonymized public code and data link at \url{https://github.com/tewiSong/MemPlan}. Appendix~\ref{app:reproducibility} states that the released materials include configuration files, PDDL construction scripts, evaluation scripts, and reproduction instructions.

\item {\bf Experimental setting/details}
    \item[] Question: Does the paper specify all the training and test details (e.g., data splits, hyperparameters, how they were chosen, type of optimizer) necessary to understand the results?
    \item[] Answer: \answerYes{}
    \item[] Justification: Section~\ref{sec:exp_setup} specifies the main experimental setup, including environments, episode counts, baselines, metrics, interface model, and planner. Appendices~\ref{app:experimental_details}--\ref{app:baseline_details} provide additional evaluation, data, transfer, baseline, output-validation, and training details, including the interface-training hyperparameters in Table~\ref{tab:interface_hyperparameters}.

\item {\bf Experiment statistical significance}
    \item[] Question: Does the paper report error bars suitably and correctly defined or other appropriate information about the statistical significance of the experiments?
    \item[] Answer: \answerYes{}
    \item[] Justification: Appendix~\ref{app:statistical_reporting} reports the statistical procedure used for the main experimental metrics. We report two-sided 95\% bootstrap confidence intervals with 10{,}000 resamples for ALFWorld and Robotouille, use episode or completed-run resampling units as appropriate, and additionally report paired differences when methods are evaluated on matched episodes; for TextWorld, we report per-instance scores and group averages because only 7 task instances are used.

\item {\bf Experiments compute resources}
    \item[] Question: For each experiment, does the paper provide sufficient information on the computer resources (type of compute workers, memory, time of execution) needed to reproduce the experiments?
    \item[] Answer: \answerYes{}
    \item[] Justification: Appendix~\ref{app:reproducibility} reports the compute resources used for training and evaluation, including CPU type, GPU type, GPU memory, allocated system memory, fine-tuning GPU-hours, evaluation wall-clock time, and the CPU/GPU division between Fast Downward planning and Qwen2.5-7B interface calls.

\item {\bf Code of ethics}
    \item[] Question: Does the research conducted in the paper conform, in every respect, with the NeurIPS Code of Ethics \url{https://neurips.cc/public/EthicsGuidelines}?
    \item[] Answer: \answerYes{}
    \item[] Justification: The research evaluates planning agents in simulated text environments and uses existing benchmark environments. The paper does not involve human-subject experiments, private data collection, or deployment on real users.

\item {\bf Broader impacts}
    \item[] Question: Does the paper discuss both potential positive societal impacts and negative societal impacts of the work performed?
    \item[] Answer: \answerYes{}
    \item[] Justification: Appendix~\ref{app:reproducibility} discusses positive impacts, including improved reliability and token efficiency for long-horizon language agents, and negative impacts, including possible failures under partial observability and misuse of autonomous planning systems.

\item {\bf Safeguards}
    \item[] Question: Does the paper describe safeguards that have been put in place for responsible release of data or models that have a high risk for misuse (e.g., pre-trained language models, image generators, or scraped datasets)?
    \item[] Answer: \answerNA{}
    \item[] Justification: The paper releases code and data for a domain-specific planning framework evaluated in simulated text environments. It does not release a high-risk pretrained language model, image generator, scraped dataset, or other asset requiring special misuse safeguards.

\item {\bf Licenses for existing assets}
    \item[] Question: Are the creators or original owners of assets (e.g., code, data, models), used in the paper, properly credited and are the license and terms of use explicitly mentioned and properly respected?
    \item[] Answer: \answerYes{}
    \item[] Justification: The paper cites the original works for the main existing assets, including TextWorld, ALFWorld, Robotouille, Fast Downward, Qwen2.5-7B, GPT-4o, ReAct, and LLM-DP. Appendix~\ref{app:reproducibility} states the license information for the used assets, including MIT licenses for TextWorld, ALFWorld, and Robotouille, GPLv3 for Fast Downward, Apache License 2.0 for Qwen2.5-7B-Instruct, and MIT License for the released MemPlan code.

\item {\bf New assets}
    \item[] Question: Are new assets introduced in the paper well documented and is the documentation provided alongside the assets?
    \item[] Answer: \answerYes{}
    \item[] Justification: The paper introduces and releases the MemPlan code and data as new assets at \url{https://github.com/tewiSong/MemPlan}. Appendix~\ref{app:reproducibility} states that the released repository includes configuration files, PDDL construction scripts, evaluation scripts, reproduction instructions, and an MIT License for the released MemPlan code.

\item {\bf Crowdsourcing and research with human subjects}
    \item[] Question: For crowdsourcing experiments and research with human subjects, does the paper include the full text of instructions given to participants and screenshots, if applicable, as well as details about compensation (if any)?
    \item[] Answer: \answerNA{}
    \item[] Justification: The paper does not involve crowdsourcing experiments or research with human subjects. The experiments are conducted in simulated benchmark environments.

\item {\bf Institutional review board (IRB) approvals or equivalent for research with human subjects}
    \item[] Question: Does the paper describe potential risks incurred by study participants, whether such risks were disclosed to the subjects, and whether Institutional Review Board (IRB) approvals (or an equivalent approval/review based on the requirements of your country or institution) were obtained?
    \item[] Answer: \answerNA{}
    \item[] Justification: The paper does not involve crowdsourcing or human-subject research, so IRB approval or equivalent review is not applicable.

\item {\bf Declaration of LLM usage}
    \item[] Question: Does the paper describe the usage of LLMs if it is an important, original, or non-standard component of the core methods in this research? Note that if the LLM is used only for writing, editing, or formatting purposes and does \emph{not} impact the core methodology, scientific rigor, or originality of the research, declaration is not required.
    \item[] Answer: \answerYes{}
    \item[] Justification: The paper explicitly uses LLMs as core method components. Section~\ref{sec:interfaces} describes the offline-trained language--symbol interface for observation-to-fact parsing and symbolic-action-to-command grounding, Section~\ref{sec:exp_setup} states that Qwen2.5-7B is fine-tuned and frozen for evaluation, and Appendix~\ref{app:interface_training} provides the interface training and validation details.

\end{enumerate}
\end{document}